\documentclass[journal]{IEEEtran}

\ifCLASSINFOpdf
\else
   \usepackage[dvips]{graphicx}
\fi
\usepackage{url}

\usepackage[rm={tabular, lining}, sf={tabular, lining},
             tt={monowidth, tabular, lining}]{cfr-lm}
\usepackage{graphicx}
\usepackage{amsmath}
\usepackage[onehalfspacing]{setspace}     
\usepackage{courier}
 
\usepackage{booktabs}
\begin{document}

\title{DeepApple: Deep Learning-based Apple Detection using a Suppression Mask R-CNN}

\author{Pengyu Chu, Zhaojian Li, Kyle Lammers, Renfu Lu, and Xiaoming Liu


}

\maketitle

\begin{abstract}
Robotic apple harvesting has received much research attention in the past few years due to growing shortage and rising  cost in labor. One key enabling technology towards automated harvesting is accurate and robust apple detection, which poses great challenges as a result of the complex orchard environment that involves varying lighting conditions and foliage/branch occlusions. This letter reports on the development of a novel deep learning-based apple detection framework named DeepApple. Specifically, we first collect a comprehensive apple orchard dataset for ‘Gala’ and ‘Blondee’ apples, using a color camera, under different lighting conditions (sunny vs. overcast and front lighting vs. back lighting).  We then develop a novel suppression Mask R-CNN for apple detection, in which a suppression branch is added to the standard Mask R-CNN to suppress non-apple features generated by the original network. Comprehensive evaluations are performed, which show that the developed suppression Mask R-CNN network outperforms state-of-the-art models with a higher F1-score of 0.905 and a detection time of 0.25 second per frame on a standard desktop computer.
\end{abstract}




\section{Introduction}
\label{sec:intro}
Fruit harvesting is highly labor-intensive and cost-heavy; it is estimated that the labor needed for apple harvesting alone is more than \textit{10 million worker hours} annually, attributing to approximately $15\%$ of the total production cost in U.S. \cite{Apple_stat}.
 Growing labor shortage and rising labor cost have steadily eroded the profitability and sustainability of the fruit industry. 
Furthermore, manual picking activities constitute great risks of  back strain and musculoskeletal pain to fruit pickers due to repetitive hand motions, awkward postures when picking fruits at high locations or deep in the canopy, and ascending and descending on ladders with heavy loads \cite{BackStrain}. Therefore, there is an imperative need for the development of robotic mass harvesting systems to tackle labor shortage, lower human injury risks, and improve productivity and profitability of the fruit industry.

The first and foremost task in robotic harvesting is apple detection, which identifies apples in the area of interest and provides targets for the robot to perform subsequent actions. Due to the low cost of cameras and the tremendous advances in computer vision  \cite{patricio2018computer}, image-based apple detection systems have gained great popularity in robotic fruit harvesting since the late 1980s. Specifically, in \cite{slaughter1987color, sites1988computer},  a simple thresholding method is developed to generate a binary image with smoothing filters that eliminate noise and irrelevant details. The large segmented regions are then recognized as fruits. This method is easy to implement but it is susceptible to varying lighting conditions. A circular Hough Transform is also proposed to obtain binary edge images along with a matrix of votes on the detection candidates \cite{whittaker1987fruit,benady1992locating}.   This approach works well with a simple background but is less applicable in a complex structured environment, such as in a dense fruit orchard. Another idea is to combine shape and texture of the fruit to generate a richer set of feature representations \cite{qiu1992maturity,cardenas1991machine,levi1988image,zhao2005tree}. By comparing the differences between fruit and leaves in texture, specific fruit or vegetable like broccoli are then detected. However, this method relies on hand-crafted features and is sensitive to lighting conditions and occlusions. 

With rapid advancements in deep learning in recent years, deep neural networks (DNNs) have  found great successes in object detection and semantic image segmentation \cite{sa2016deepfruits, bargoti2017image}. DNN-based methods can learn feature representations automatically without the need of feature hand-engineering. For example, Dias et. al.\cite{dias2018apple}  used a combination of convolutional neural network (CNN) and support vector machine (SVM) to extract features of apple blossoms in a complex background, which shows a good performance of $82.2\%$ F1-score. More recently, region-based convolutional neural network (R-CNN) has gained great popularity in object detection \cite{girshick2015region}. R-CNN utilizes regions of interest produced by selective search \cite{uijlings2013selective} and then regresses bounding box location with classification. 
 A further extension, called Mask R-CNN \cite{he2017mask}, enhances end-to-end classification and segmentation capacities with a region proposal network and an additional mask branch network, which has achieved  state-of-the-art results on varioius datasets like PASCAL-VOC  \cite{everingham2010pascal}. The application of  R-CNN on apple detection \cite{kang2019fruit} reaches an F1-score of $0.813$ while Mask R-CNN for orange detection \cite{ganesh2019deep} reaches an F1-score of $0.89$. Different from the above two-stage networks, one-stage detection network like YOLOv3 \cite{redmon2018yolov3} also achieves a decent performance (an F1-score of 0.817) with reduced computational complexity. 

Despite the aforementioned developments, accurate apple perception to support robotic harvesting in real orchard environments remains a great challenge. Existing methods either provide insufficient accuracy \cite{redmon2018yolov3,kang2019fruit} or are based on simple structured orchards with little occlusion and stable lighting conditions \cite{dias2018apple, bulanon2010fruit}. As such, the goal of this study is to develop a robust and accurate apple detection framework to support robotic harvesting in real orchard environment. Towards this end, we collected a comprehensive dataset from two commercial orchards for two varieties of apples with distinct colors under various lighting conditions. Furthermore, we extended the well-known Mask R-CNN \cite{he2017mask} with a suppression network, hereinafter referred to as suppression Mask R-CNN, to improve detection performance. Performance evaluations for apple detection were then conducted to compare the proposed suppression Mask R-CNN with state-of-the-art models. 

The contributions of this study are summarized as follows:

1) We collect and process a comprehensive orchard dataset with multiple apple varieties under various lighting conditions in real orchard environment.

2) We develop a new deep network, suppression Mask R-CNN, to remove false detection due to occlusion and thus increase the accuracy and robustness of apple detection.

3) Extensive evaluations show that the proposed suppression Mask R-CNN achieves state-of-the-art performance.

The remainder of this paper is organized as follows. Section.~\ref{sec:datapre} presents the orchard data collection and processing. The suppression Mask R-CNN is then detailed in Section.~\ref{sec:method}. Experiments are performed in Section.~\ref{sec:exp} to evaluate the proposed framework with comparisons to state-of-art benchmarks. Finally, Section.~\ref{sec:con} concludes the paper with discussions on future work.

\section{Data Collection and Processing}
\label{sec:datapre}
\begin{figure}[!h]
\centerline{\includegraphics[width=0.9\columnwidth]{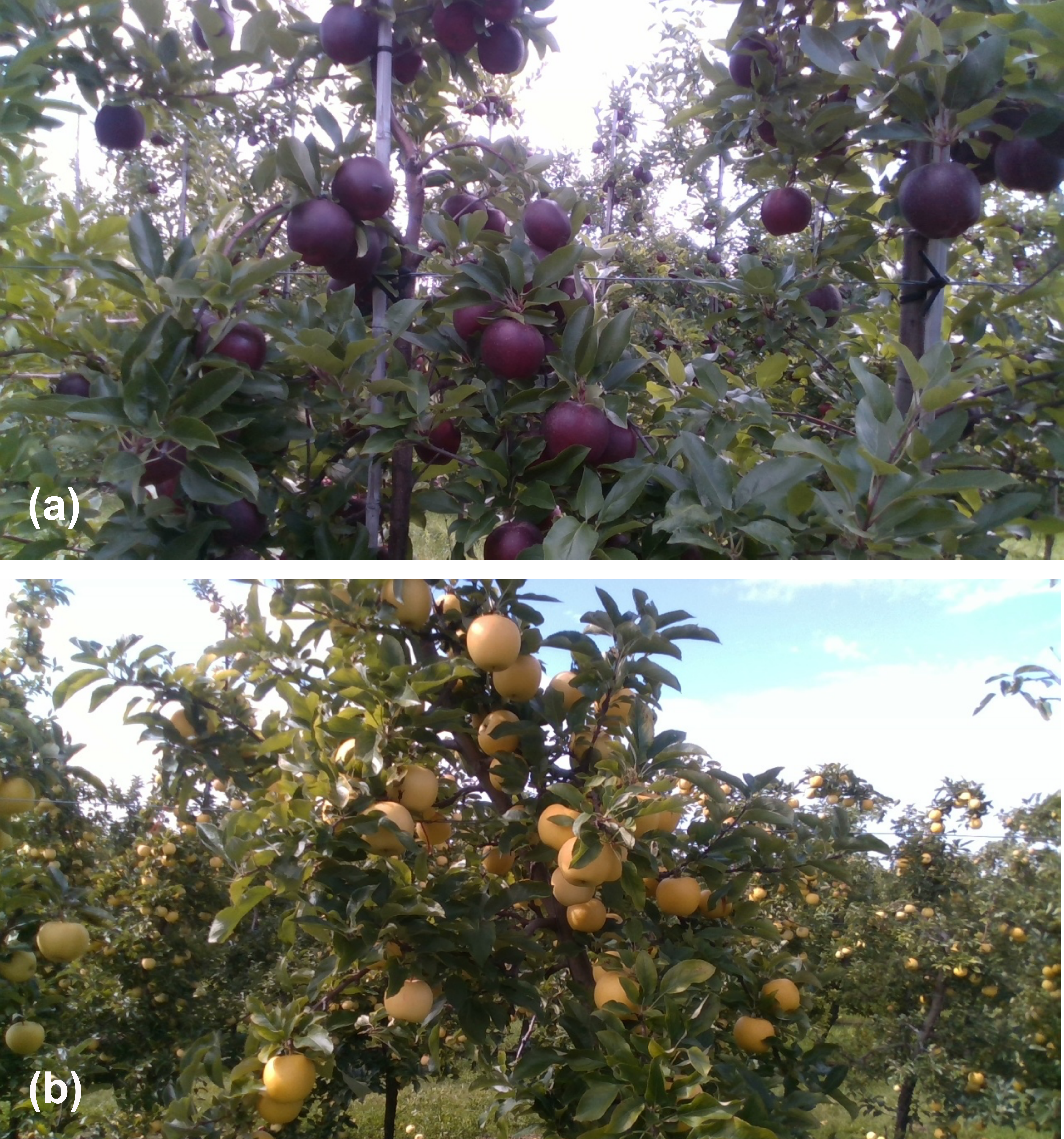}}
\caption{Two sample images from the collected dataset: (a) a sample image of Gala orchard; (b) a sample image of Blondee orchard.}
\label{fig:apple}
\end{figure}

In this study, apple images of ‘Gala’ and ‘Blondee’ varieties were taken in two commercial orchards in Sparta, Michigan, USA during the 2019 harvest season. The two apple varieties have distinct color characteristics; ‘Gala’ apples are red over a yellow background, while ‘Blondee’ apples have a smooth yellow skin (see Fig.~ \ref{fig:apple}). A RGB camera with a resolution of $1,280$x$720$ was used to take images of apples at a distance of $1\sim2$ meters to the tree trunk, which is the typical range of harvesting robots \cite{de2011design}. The images were collected across multiple days to cover both cloudy and sunny weather conditions. In a single day, the data were also collected at different times of the day, including 9:00am in the morning, noon, and 3:00pm in the afternoon, to cover different lighting angles: front-lighting, back-lighting, side-lighting, and scattered lighting. When capturing images, the camera was placed parallel to the ground and directly facing the trees to mimic the harvesting scenario. A total of $1,500$ images were captured where two sample images are shown in Fig.~\ref{fig:apple}. 

We next processed the collected raw orchard images into formats that can be used to train and evaluate deep networks. Specifically, apples in the images were annotated by rectangles using VGG Image Annotator \cite{dutta2019vgg} and the annotation was then compiled into the human-readable format. Compared to polygon and mask annotations, rectangular annotation used here accelerates data preparation, particularly in dense images like our dataset. The annotated dataset was then split into training, validation, and test subsets with the apple quantities of $10,530$, $4,203$, and $4,795$, respectively.

\begin{figure*}
\centerline{\includegraphics[width=\textwidth,height=8.3cm]{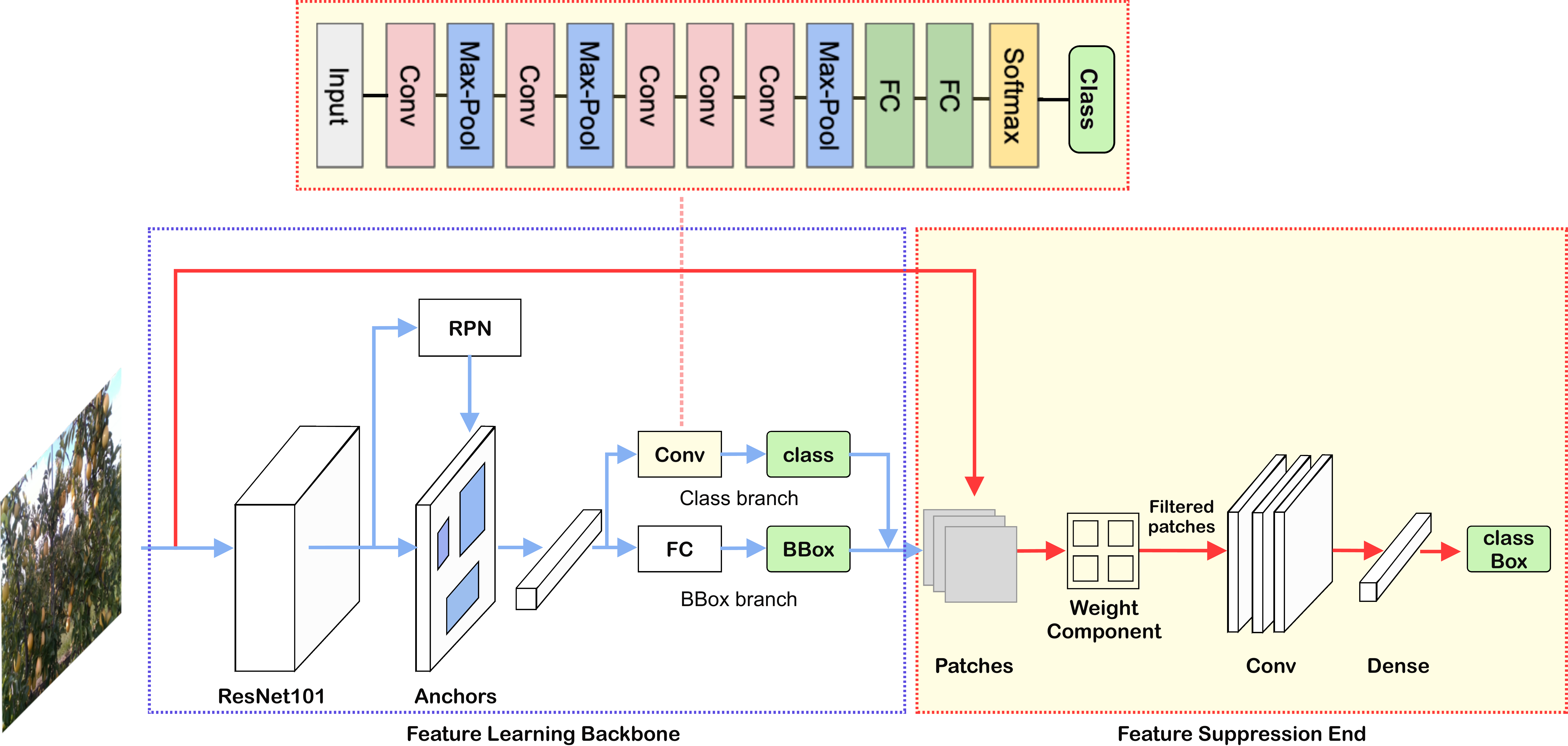}}
\caption{Structure of the suppression Mask R-CNN. It consists of a feature learning backbone and a feature suppression end. The feature learning backbone is a deep network to learn apple features while the feature suppression end, consisting of a weighting component and a shallow ConvNet, is used to filter non-apple regions.}
\label{fig:cnn}
\end{figure*}

\section{Suppression Mask R-CNN}
\label{sec:method}
This section describes the development of a new deep learning-based apple detection approach that systematically combines a DNN backbone and a RGB feature-based suppression network. As shown in Fig.~\ref{fig:cnn}, the proposed suppression Mask R-CNN consists of two parts: a feature learning backbone from Mask R-CNN \cite{he2017mask} and a feature suppression end. The former is used to learn apple features and generate region proposals. In the meantime, due to the foliage and branch occlusions, it will also learn foliage and branch features that can cause false detection. As such, we introduce a suppression network to filter non-apple features to improve detection performance by exploiting a combination of clustered features and convoluted features. These two networks are trained separately to avoid generating similar feature maps. We next discuss the two networks in more details.

\subsection{Feature Learning Backbone}

The feature learning network uses the Mask R-CNN backbone \cite{he2017mask} and follows Mask R-CNN's two-stage learning procedures with two modifications. First, the convolutional backbone in Mask R-CNN is used for feature extraction over an entire image, and is applied as the network backbone for bounding-box recognition. In this study, we instantiate feature learning backbone with ResNet-101-FPN \cite{he2017mask} as its backbone. ResNet101 outperforms other single ConvNet mainly  because it maintains strong semantic features at various resolution scales. Even though ResNet101 is a deep network, the residual blocks and dropouts function help it avoid gradient vanishing and exploding problems. Then similar to \cite{he2017mask}, we use a Region Proposal Network (RPN) \cite{ren2015faster} to generate object regions. RPN is 
a small convolutional network which can convert feature maps into scored region proposals around where the object lies. These proposals with certain height and width are called anchors, which  are a set of predefined bounding boxes. The anchors are designed to capture the scale and aspect ratio of specific object classes and are typically determined based on object sizes in the dataset. In the second stage,  class and box offset are predicted by virtue of Faster R-CNN \cite{ren2015faster} that applies bounding box classification and regression in parallel. As shown in Fig.~\ref{fig:cnn}, another  network is employed to take the proposed regions from the first stage and assign them to  specific areas of a feature map obtained at the second stage. After scanning these areas, the network generates object classes and bounding boxes simultaneously \cite{he2017mask}.

Second, for improving the recall or true detection of our algorithm, we introduce a convolutional structure (as shown in Fig.~\ref{fig:cnn}) in the class branch to learn additional feature representations. The features condensed from the Mask R-CNN backbone and fully connected layers may have lost considerable details of apples. Since images have many occlusions in our dataset, the deep network can treat some partial foliage features as apple features. These additional feature representations will enable the  identification of certain regions in an image as an occluded apple or foliage. Furthermore, we freeze the layers in the ResNet101 backbone and train this class branch independently in case there are many overlaps compared to our main network.

\subsection{Feature Suppression End}
After the feature learning step, bounding boxes of apple candidates are obtained. The image patches inside the bounding boxes are then fed into a feature suppression end to remove mis-labeled candidates. Since the feature learning backbone may have learned wrong inference features like leaves with apple-like shapes, the purpose of this suppression network is to avoid that non-apple regions flow into the last decision layer. 

\begin{figure}
\centerline{\includegraphics[width=\columnwidth]{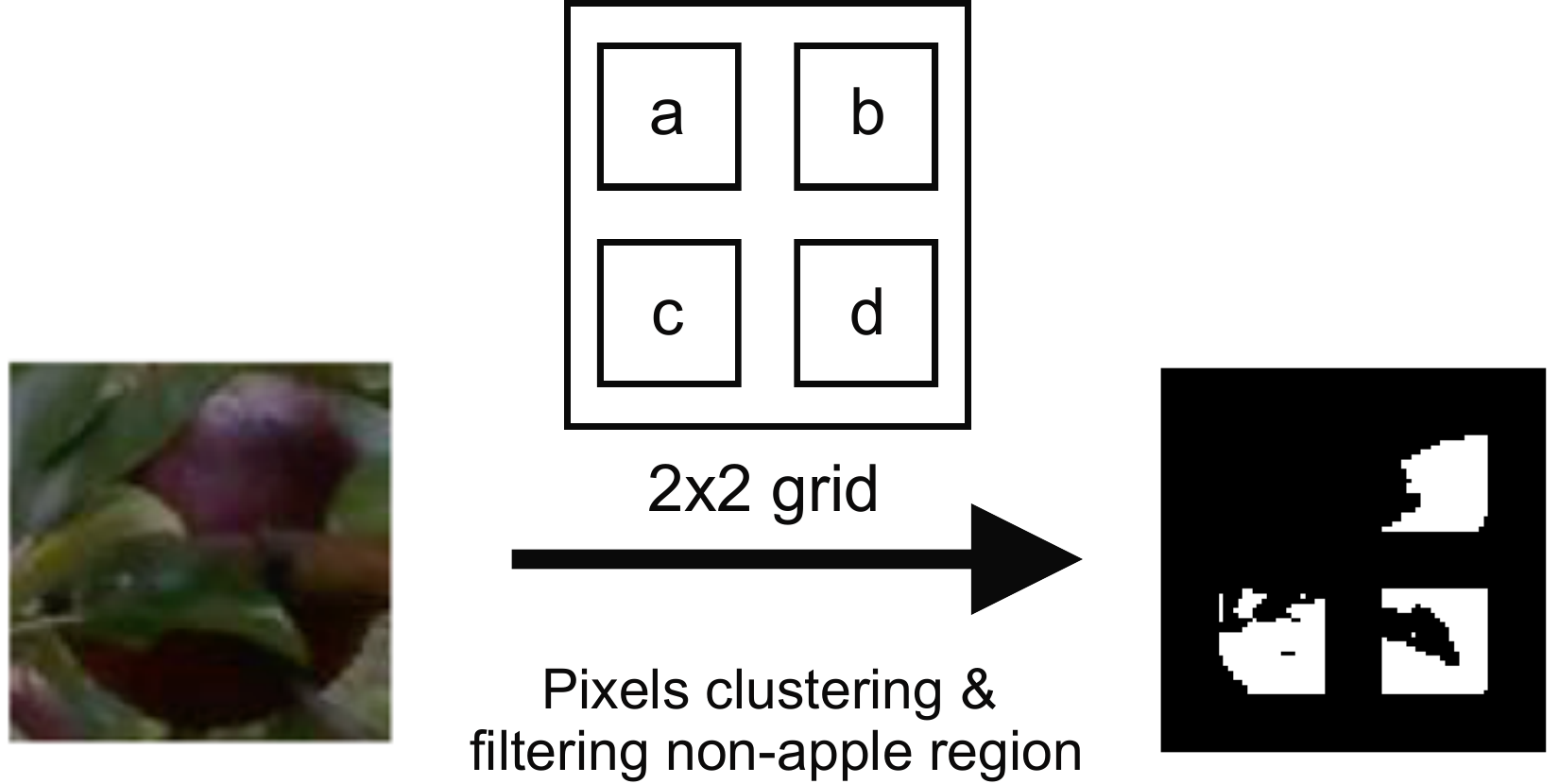}}
\caption{Diagram of the weighting scheme. The sliced image in a bounding box is first partitioned to 2x2 grids ($a, b, c, d$). Based on clustering and pixel counting, apple pixels are identified and set as 1 (white in the right image) while other pixels are set as 0 (black in the right image).}
\label{fig:cluster}
\end{figure}

Specifically, the suppression network consists of a weighting component and a shallow ConvNet. The weighting component is a 2x2 grid clustering layer that aims to determine apple regions in terms of apple pixel counts. The motivation is that in our annotated dataset, each apple is annotated in the center of a bounding box and occupies the major area in that bounding box. Even though the canopies always partially occlude the apple, the pixels corresponding to  the apple 
are still in the majority. Therefore, as shown in Fig.~\ref{fig:cluster}, we divide each bounding box in the training dataset into four regions, $a, b, c, d,$ as a 2x2 grid. Furthermore, we use K-means clustering \cite{krishna1999genetic} to group similar pixels and obtain several clusters. After clustering, we label each pixel with its class number $i$, $i=1,2,3, ... n$, with $n$ being the pre-specified cluster numbers (In our experiments, we use $n=3$). Since the class associated with the most pixels will correspond to the apple region, we select the ``apple" region from the four grids and define its pixel counts as $N^a,\, N^b,\, N^c,\,$ and $N^d$, respectively.  We will then set the apple region pixels as 1 whereas other pixels are assigned to zero. A sample output is shown in Fig.~\ref{fig:cluster}.
The weighting component keeps the objective information and generates an output with only apple pixels, which makes it more efficient to train feature suppression network that we will discuss later. The other merit of weighting component is that if the previous network recognizes a leaf as an apple, only leaf pixels are treated as objectives and flow to next ConvNets. That makes suppression network easy to discriminate apple and non-apple objectives.

The second component is a shallow convolutional network that is used to learn apple features based on filtered patches generated by the weighting component. Compared to the feature learning backbone, the features to learn in this shallow network is less. Only three convolution layers (3x3x32, 3x3x32, 3x3x64) associated with pooling layers (17x17x32, 7x7x32, 2x2x64) and ReLU as activation are used to fit the discrimination function. Two additional dense layers are employed to flatten feature maps and produce decision. This network has a total of $45,153$ trainable parameters. The detailed architecture is described in Fig.~\ref{fig:cnn}. With the help of feature suppression end, we suppress non-apple class flowing into the decision layer and it does not significantly increase inference time since the depth of the feature suppression end is small. The proposed feature suppression end can be viewed as a filter to efficiently reduce false alarms. 

\subsection{Loss Functions}
Since we train the feature learning backbone and the suppression network separately, we define two loss functions as follows.
For the feature learning backbone, we use the same loss function with Mask R-CNN \cite{he2017mask}, which defines a multi-task loss on each sampled region of interest as $L_{backbone} = L_{cls} + L_{box}$, where $L_{cls}$ and $L_{box}$ are, respectively, classification loss and bounding box loss defined as: 

\begin{equation}
L_{backbone} = \frac{1}{N_{cls}}\Sigma_i L_{cls}(p_i, p_i^*) + \frac{\lambda}{N_{box}} \Sigma_i p_i^*\cdot L_{box}(t_i, t_i^*)\\
\end{equation}
\begin{equation}
   L_{cls}(p_i, p_i^*) = -p_i^*\log p_i - (1-p_i^*)\log (1-p_i)
\label{eqn:loss}
\end{equation}
where $p_i$ and $p_i^*$ are, respectively, the predicted probability and ground truth of anchor $i$; $t_i$ and $t_i^*$ are, respectively, predicted coordinates and ground-truth coordinates; $N_{cls}$ and $N_{box}$ are normalization terms of batch size and number of anchor locations;  the loss function $L_{box}$ is the L1-smooth function \cite{girshick2015fast}, and $\lambda$ is a balancing parameter. 

For feature suppression end, we define $L_{end}$ as the average binary cross-entropy loss. For a patch associated with ground-truth class, $L_{end}$ is defined as:

\begin{equation}
    L_{end} = - [y\log{\hat{y}}+(1-y)\log(1-\hat{y})]
\label{eq:4}
\end{equation}
where $y$ is the ground truth and $\hat{y}$ is the prediction.

\section{Experiment Results}
\label{sec:exp}
\begin{figure}
\centerline{\includegraphics[width=\columnwidth]{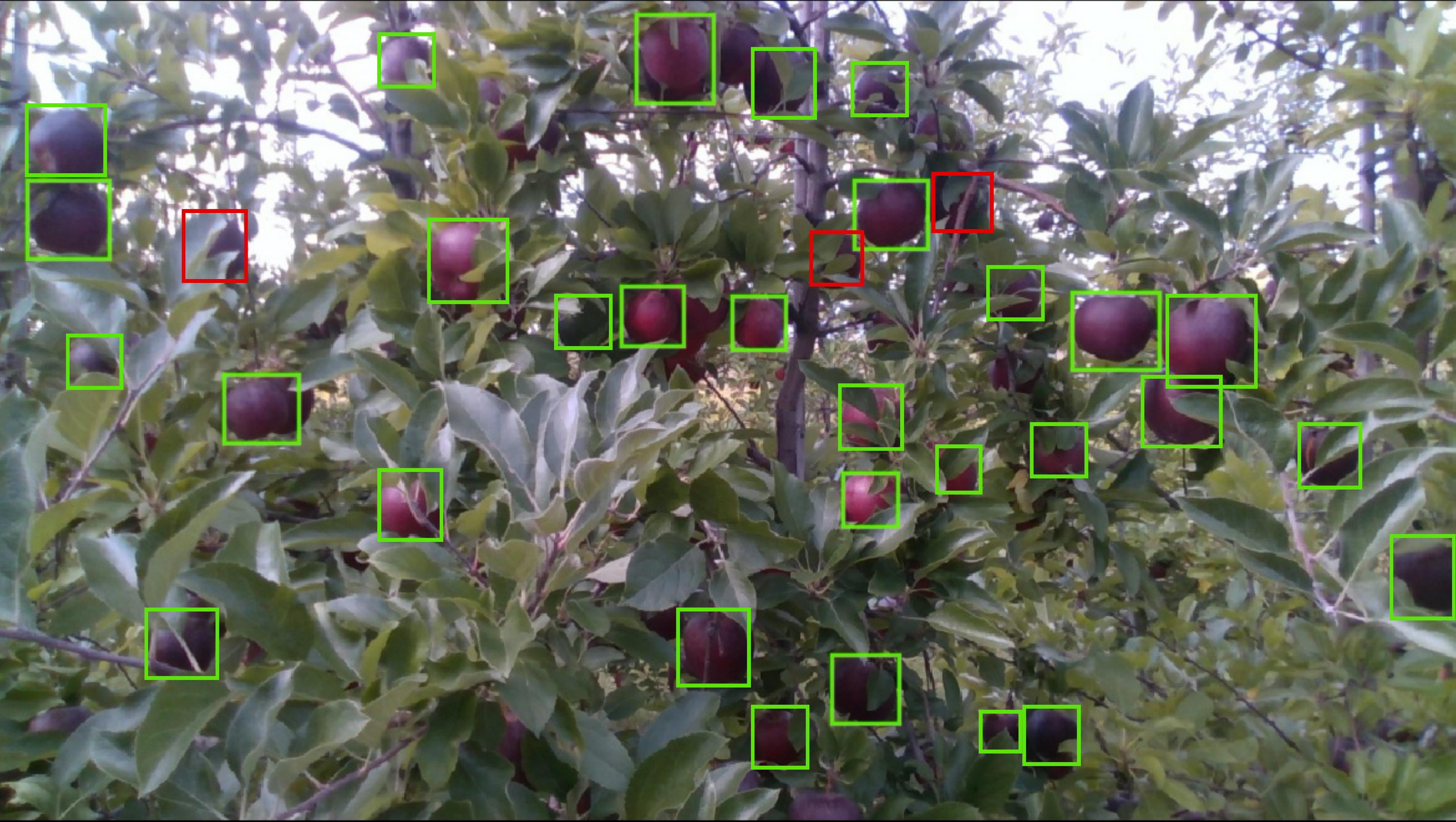}}
\caption{An example of Gala apple detection using our suppression Mask R-CNN. It shows that the majority of apples are detected (green bounding boxes) but there are still $3$ apples missed (red bounding boxes) due to heavy occlusion.}
\label{fig:detection}
\end{figure}

In this section, we evaluate the efficacy of the suppression Mask R-CNN with the processed data as discussed in Section.~\ref{sec:datapre}. The network hyper-parameters, including the momentum, learning rate, decay factor, training steps, and batch size, are set as $0.9$, $0.001$, $0.0005$, $934$, and $1$, respectively, through cross-validation. The input image size is $1,280$x$720$, which is aligned with the camera resolution. To better analyze the training process, we set up $100$ epochs for training. We exploit a pre-trained model on COCO dataset \cite{lin2014microsoft} to warm-start the training process and it generally only needs $50$ epochs to converge. A detection example  is shown in Fig.~\ref{fig:detection}, where green boxes represent correctly identified apples while red boxes represent missed detection.

\begin{table}[!h]
    \setlength{\tabcolsep}{3pt}
    \centering
    \caption{Performance comparison between state-of-art networks and our proposed Suppression Mask R-CNN with two parameter configurations ($C_1$ and $C_2$).}
    \begin{tabular}{c|c|c|c}
         & Precision & Recall & F1-score \\
        \hline
        Faster R-CNN & 0.761 & 0.889 & 0.820 \\
        \hline
        Mask R-CNN(ResNet50) & 0.753 & 0.892 &  0.817\\
        \hline
        Mask R-CNN(ResNet101) & 0.789 & 0.927 & 0.852 \\
        \hline
        Mask R-CNN(ResNet152) & 0.798 & 0.928 & 0.858 \\
        \hline
        Suppression Mask R-CNN($C_1$) & \textbf{0.880} & 0.931 & \textbf{0.905} \\
        \hline
        Suppression Mask R-CNN($C_2$) & 0.801 & \textbf{0.939} & 0.864 \\
    \end{tabular}
    \label{tab:my_label}
    
\end{table}

\begin{table*}[!tp]
\centering
\caption{Performance evaluation on subset of the data with different apple varieties as well as different lighting conditions.It can be seen that similar performance are obtained in Gala and Blondee apples while back lighting can slightly decrease the performance. }
\begin{tabular}{ccccccc}
\multicolumn{6}{l}{}\\
\toprule
   & \multicolumn{6}{c}{Dataset}\\
  & \multicolumn{2}{c}{Category} & 
 \multicolumn{3}{c}{Lighting Condition} & Total \\
 \cmidrule(lr){2-3}\cmidrule(l){4-6}
 & 
 Gala & Blondee &  
 Overcast & Direct Lighting & Back Lighting & \\
 \midrule
 Number & 3,357 & 1,438 & 3,356 & 959 &480 &4,795 \\
 \midrule
    Precision & .87 & .89 & .89 & .89 & .84 & .88\\
    Recall & .93 & .93 & .93 & .93 & .93 & .93\\
    F1-score & .90 & .91 & .91 & .91 & .88 & .91\\
   \bottomrule
\end{tabular}
\label{tab:subdata_result}
\end{table*} 

To quantitatively evaluate the detection performance, we use performance metrics including precision, recall and F1-score for algorithm evaluation. All detection outcomes are divided into four types: true positive (TP), false positive (FP), true negative (TN), and false negative (FN), based on the relation between the true class and predicted class. Then precision (P) and recall (R) are defined as follows: 

\begin{equation}
    P = \frac{TP}{TP+FP} \\
    R = \frac{TP}{TP+FN}
\end{equation}

Then F1-score is defined based on precision and recall as follows:

\begin{equation}
    F1 = \frac{2\cdot P\cdot R}{P+R}
\end{equation}

To compare the performance with Mask R-CNN baselines, we first train our suppression Mask R-CNN and the state-of-art Mask R-CNNs (with ResNet50, ResNet101 and ResNet152). The Mask R-CNN with the backbone of ResNet152 has more parameters than our suppression Mask R-CNN.
Furthermore, the trained models are evaluated on the test data, whose performance are reported in Table.~\ref{tab:my_label}. It can be seen that our model obtains the best recall and F1-score, which demonstrates the effectiveness of the suppression network. In addition, the result shows that our model is better than the deeper model ResNet152.

\begin{figure}
\centerline{\includegraphics[width=\columnwidth]{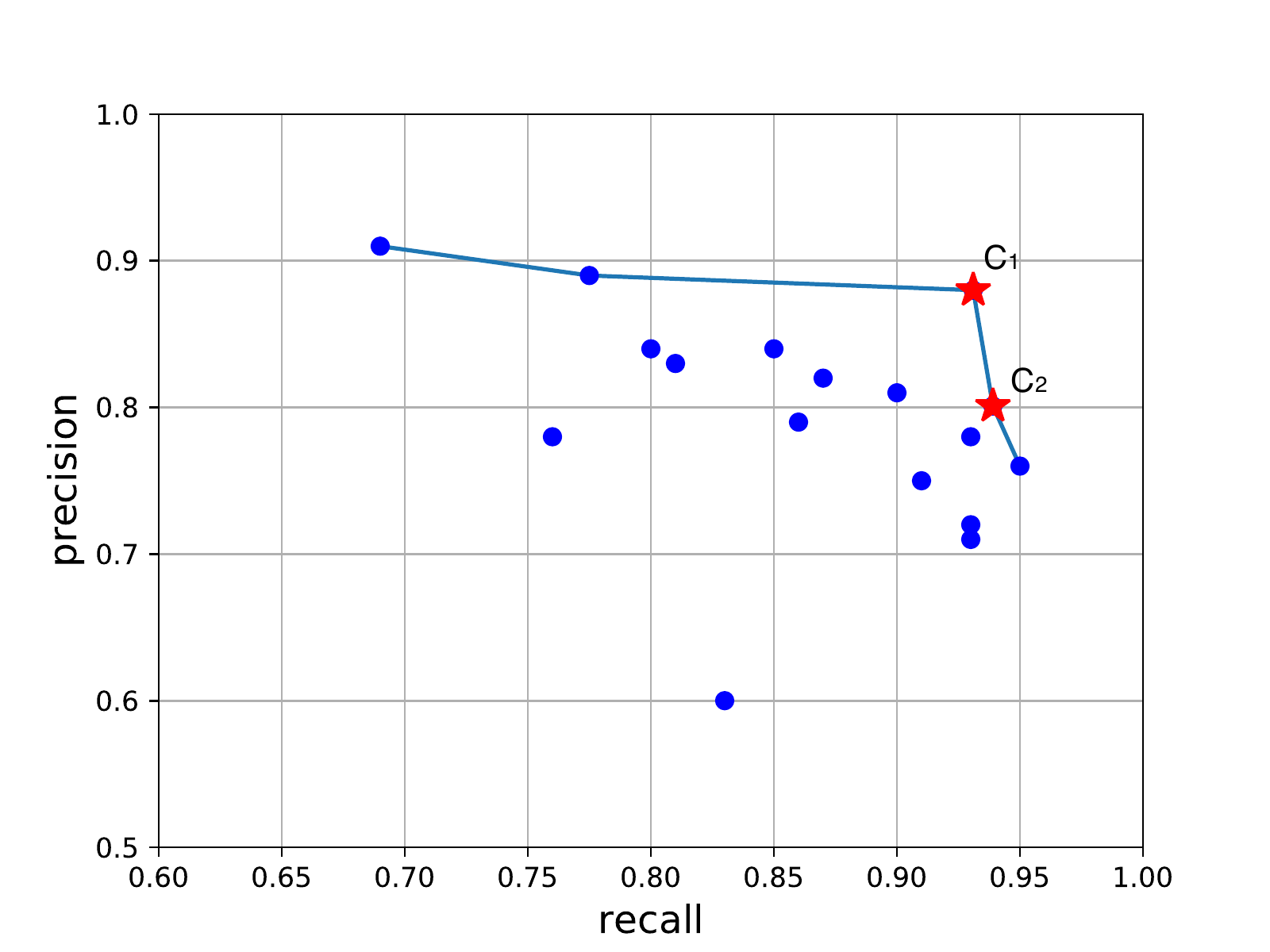}}
\caption{The Pareto plot of  recall-precision  on different combinations of $th_1$ and $th_2$.The Pareto front is shown in blue solid lines and the two configurations used to compare with the state-of-art networks (see Table~\ref{tab:my_label}) are shown in red stars.}
\label{fig:thres1}
\end{figure}

Note that the suppression network offers a tradeoff between recall and precision, that is, aggressive suppression will lead to higher precision but lower recall rate.  This tradeoff can be controlled by adjusting two confidence thresholds $th_1$ in the class branch network and $th_2$ in the feature suppression end. Then we tune both confidence thresholds during the inference process to obtain the best recall and precision of our entire model.  Fig.~\ref{fig:thres1} shows the Pareto plot, where each point represents the performance of a combination of $th_1$ and $th_2$. From the Pareto front (blue solid lines) in Fig.~\ref{fig:thres1}, we choose two ``best'' configurations $C_1$ and $C_2$, among which $C_1$ represents a better F1-score $0.905$ whereas $C_2$ achieves a better of recall rate of $0.939$.  The  detection performance is listed in Table.~\ref{tab:my_label}, which shows with $C_1$ $10\%$ increase in precision and $0.4\%$ increase in recall are achieved whereas $1.6\%$ increase in precision and $1.3\%$ increase in recall are achieved with configuration $C_2$. These results demonstrate that in both cases, our integrated class branch and the suppression end approach improve the true detection and the $C_2$ configuration significantly reduce false fruit detection rates.

In addition, we also evaluate our model in different sub-datasets. Specifically, we separate the whole dataset into several sub-datasets based on apple variety and lighting conditions. The results are summarized in the Table.~\ref{tab:subdata_result}.The results show that our model has a better performance for Blondee apples than for Gala. Compared to back lighting conditions, the detection of our model reaches a higher precision under overcast or direct lighting conditions, which indicates that artificial lighting may be helpful for further improving the performance and it will be investigated in our future work.

\section{Conclusion}
\label{sec:con}
In this study, we collected a comprehensive apple dataset for two varieties of apples with distinct yellow and red colors under different lighting conditions from the real orchard environment. A novel suppression Mask R-CNN was developed to robustly detect apples from the dataset. Our developed feature suppression network significantly reduced false detection by filtering non-apple features learned from the feature learning backbone. Our suppression Mask R-CNN demonstrated superior performance, compared to state-of-the-art models in experimental evaluations.

Our future work will include the incorporation of depth information in the network design to further improve the detection performance.  Furthermore, foliage and branches detection will be developed to provide necessary contextual information for the robot to maneuver, e.g., avoiding colliding with tree branches. Lastly, we will also investigate whether artificial lighting augmentation can enhance the detection performance.


\clearpage
\bibliographystyle{IEEEtran}
\bibliography{main}

\end{document}